\documentclass[10pt,twocolumn,letterpaper]{article}

\usepackage{wacv}
\usepackage{times}
\usepackage{epsfig}
\usepackage{graphicx}
\usepackage{amsmath}
\usepackage{amssymb}

\def\wacvPaperID{1252} 

\wacvfinalcopy 

\ifwacvfinal
\def\assignedStartPage{9876} 
\fi

\ifwacvfinal
\usepackage[breaklinks=true,bookmarks=false]{hyperref}
\else
\usepackage[pagebackref=true,breaklinks=true,colorlinks,bookmarks=false]{hyperref}
\fi

\ifwacvfinal
\else
\fi

\usepackage{Inputs/packages}

\begin{document}

\title{Zero-Pair Image to Image Translation using Domain Conditional Normalization}

\author{Samarth Shukla\textsuperscript{1} \qquad Andr\'es Romero\textsuperscript{1} \qquad Luc Van Gool\textsuperscript{1,2} \qquad Radu Timofte\textsuperscript{1} \\
\textsuperscript{1}Computer Vision Lab, ETH Z\"urich \qquad \textsuperscript{2}KU Leuven\\
{\tt\small \{samarth.shukla, roandres, vangool, timofter\}@vision.ee.ethz.ch}
}


\maketitle

\begin{abstract}
In this paper, we propose an approach based on domain conditional normalization (DCN) for zero-pair image-to-image translation, i.e., translating between two domains which have no paired training data available but each have paired training data with a third domain. We employ a single generator which has an encoder-decoder structure and analyze different implementations of domain conditional normalization to obtain the desired target domain output. The validation benchmark uses RGB-depth pairs and RGB-semantic pairs for training and compares performance for the depth-semantic translation task. The proposed approaches improve in qualitative and quantitative terms over the compared methods, while using much fewer parameters. Code available at: \url{https://github.com/samarthshukla/dcn}


\end{abstract}

\section{Introduction}
\label{sec:intro}
In recent years, image-to-image (I2I) translation has gained increasing attention thanks to Generative Adversarial Networks (GANs)~\cite{goodfellow2014generative}, and their power to learn diverse statistical distributions. The capacity of performing a cross-domain mapping between two images is applied to numerous computer vision problems such as super-resolution~\cite{wang2018esrgan,buhler2020deepsee}, colorization~\cite{zhang2016colorful}, enhancement~\cite{ignatov2017dslr}, manipulation~\cite{ntavelis2020sesame}, among others. 

Most of the current I2I techniques for paired training data~\cite{isola2017pix2pix,wang2018high} rely on the assumption that there exists a direct link between the source and target domains, whereas methods for unpaired training data~\cite{zhu2017cyclegan,lee2018drit,choi2017stargan,ma2018exemplarmunit2} assume that there does not exist any direct or indirect link between source and target domains. Some works~\cite{shukla2019extremely} assume that a couple of training pairs are available. Therefore, if we have access to disjoint sets of paired data $A \leftrightarrow B$ and $A \leftrightarrow C$, both these techniques cannot utilize this information to learn the mapping $B \leftrightarrow C$. Such a setting, where a translation between two domains that have no direct link, but an indirect connection exists is referred to as zero-pair I2I problem. Wang~\etal~\cite{Wang_2018_CVPR} demonstrated that this indirect mapping can be learned without requiring any explicit paired samples, by training independent encoders and decoders for each domain and performing feature alignment for paired samples from different domains when paired data exists. During inference, the output is obtained by composing the encoders and decoders of the corresponding source and target domains. Their approach allows large transformations among domains, but limits the scalability and compromises the computational resources, since a new encoder and decoder is required for each additional domain.

We propose a single encoder-decoder network for translation from any source domain to any target domain, thus utilizing fewer parameters to produce similar or better results. We achieve this through domain conditional normalization in the decoder. For such an approach to be successful in the zero-paired setting, we argue that it must satisfy either of the two conditions: 1) the latent space representation must be aligned (domain invariant) for all the involved domains and the decoder must couple conditioned information about the output domain, or 2) the decoder must exploit conditional information representing both input and output domains, and learn to map samples for all desired domain pairs. We propose a unique approach for each of these two conditions. Our main contributions are as follows:

\begin{enumerate}
\vspace{-0.3cm}
\itemsep0em 
    \item[(i)] We introduce two different approaches using domain conditional normalization, which can perform any-to-any domain mapping using a single encoder-decoder architecture.
    \item[(ii)] We compare our method with the state-of-the-art for zero-pair I2I translation problem, and show that our method performs better while using a much lower number of parameters.
    \vspace{-0.3cm}
\end{enumerate}

\begin{figure}[t]
  \centering
  \begin{subfigure}[b]{0.45\textwidth}
    \centering
    \includegraphics[width=\textwidth]{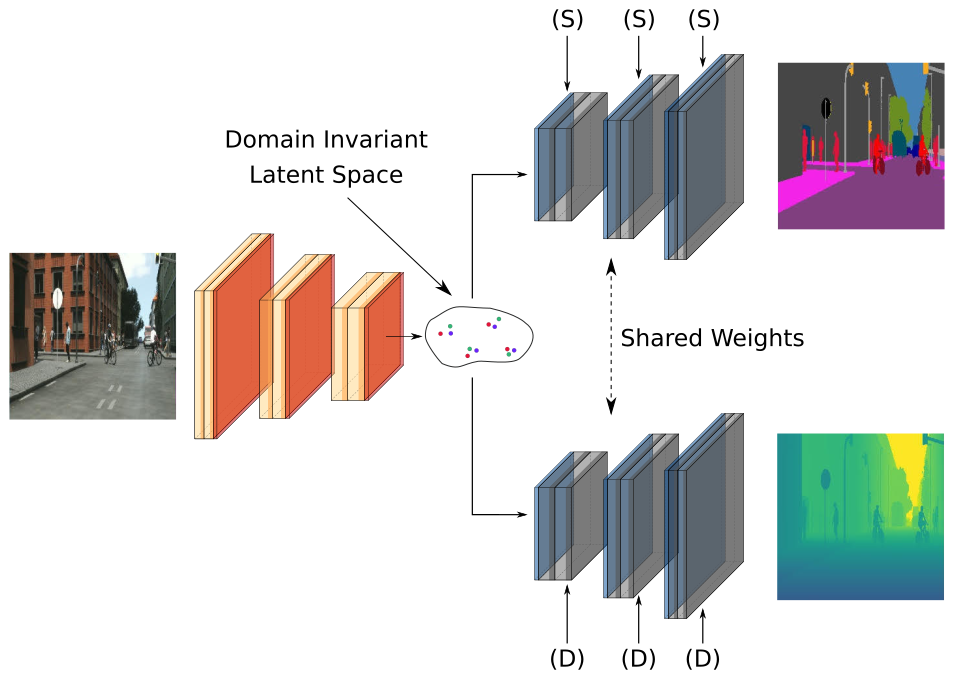}
    \caption{DCN-0}
    \label{fig:approach_1}
  \end{subfigure}
  \hfill
  \begin{subfigure}[b]{0.45\textwidth}
    \centering
    \includegraphics[width=\textwidth]{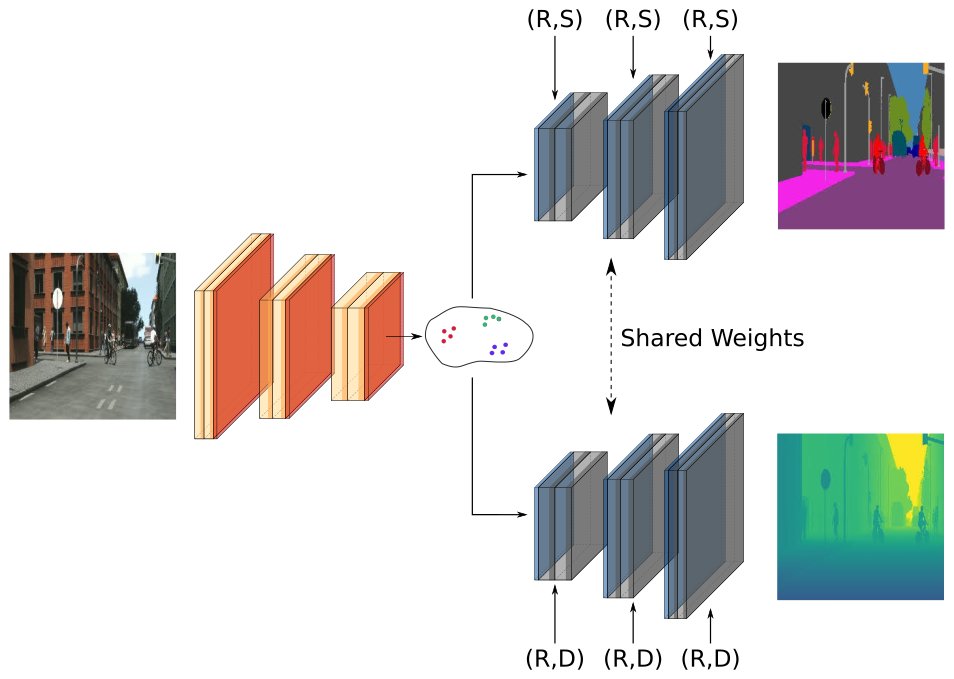}
    \caption{DCN}
    \label{fig:approach_2}
  \end{subfigure}
  \caption{Our two proposed domain conditional normalization solutions. (a) The first approach, DCN-0 uses a domain invariant latent space as input to the decoder, where normalization is performed conditioned on the desired target domain. (b) In the second approach, DCN does not impose any constraint on the latent space, and the decoder produces the desired output through conditional normalization imposed on source-target domain pairs. Symbols $R$, $D$, and $S$ represent the RGB, depth, and semantic segmentation domains, respectively.}
  \label{fig:overview}
\end{figure}

In summary, our method tries to achieve a good mapping between unseen domain pairs while using a single network, accomplished through domain conditional batch normalization. We experimentally evaluate our method for the problem of zero-pair I2I translation across unseen domain pair of depth and semantic segmentation, where we assume we have access to RGB-Depth and RGB-Semantics paired training data. An overview of our two approaches is shown in Figure~\ref{fig:overview}.

\section{Prior Work}
\label{sec:related_work}

\subsection{Multi-domain I2I translation}
Generally, I2I translation models rely on an encoder-decoder architecture, where the encoder serves as feature encoding and the decoder as a conditional generator. Depending on the number of domains and how big the domain gap is, the system could be modeled using one architecture per domain or a single multi-domain architecture. Particularly, CycleGAN~\cite{zhu2017cyclegan}-based approaches~\cite{lee2018drit,huang2018munit,zhu2017bicycle} are suitable for a small number of domains, and they allow large gaps in the cross-domain mapping thanks to the specialized uni-domain system (\eg, semantic-layout inpainting, anime to portrait, etc). Conversely, approaches based on StarGAN~\cite{choi2017stargan,pumarola2018ganimation,romero2019smit} study multiple domains for small transformations, where the domain gap is rather small (\eg, facial attributes or weather condition as domains, etc.). However, applications using single encoder-decoder multi-domain I2I translation models for domains with large gaps remain relatively unexplored.

\subsection{Conditional Normalization}

Conditional normalization is performed on neural network layer activations by first subtracting their mean and then, dividing by the variance \ie normalizing the layer activations. Therefore, the normalized output is transformed through a learned affine transformation, which its parameters depends on the conditioned input. There are different conditional normalization layers in the literature depending on how this affine transformation is defined, such as Conditional Batch Normalization (CBN)~\cite{de2017modulating} and Adaptive Instance Normalization~\cite{huang2017arbitraryadain} (AdaIN). Conditional normalization methods help in incorporating additional information into the models. CBN is suitable for injecting discrete information, whereas AdaIN works better for a continuous representation. Since we only have to use information that identifies the input and output domains we require, we adopt CBN for our proposed method. Having the input and output domains represented through the normalization layers alleviates the need to have separate domain specific encoders and decoders and enables us to perform cross domain transformations between all domain pairs using only a single encoder and decoder.

\subsection{Domain Invariance}
Latent space invariance across different domains can be seen as a problem of domain adaptation. Ganin~\etal~\cite{ganin2016domain} first introduced a domain adversarial loss in order to get latent space features which cannot be distinguished for the different domains. Wang~\etal~\cite{Wang_2018_CVPR} use different encoders and decoders per domain, and perform the zero-pair translation task by minimizing with a feature matching loss the output of the encoder for paired samples from different domains. We differ from them in two ways. First, we only use a single encoder-decoder network to generate samples for all domains, thus significantly reducing the number of parameters of the network and demonstrate that this joint representation achieves significantly better results. Second, we show that the zero-pair translation can be performed without explicitly enforcing a domain invariant representation and instead using an input-output domain conditional normalization, and show such an approach also leads to remarkably improved performance while having minimal extra model parameters.

\section{Method}
\label{sec:method}

In this section, we describe the zero-paired I2I translation problem and the different approaches to conditional normalization we implemented in order to improve model performance.

Let $\mathcal{X}^{1}, \mathcal{X}^{2}, ..., \mathcal{X}^{n}$ denote the set of $N$ domains. We wish to perform a multi-domain I2I translation between any two domains $\mathcal{X}^{j}$ and $\mathcal{X}^{k}$. Our final objective is to learn a mapping function $\mathbb{F}$ that allows us to perform multi-domain transformations by using an encoder ($\mathbb{E}$) and a conditional decoder ($\mathbb{G}$) to translate an input image from a source domain to an output image corresponding to a target domain.

We primarily focus on the problem of zero-pair I2I translation, where we translate between domains where direct paired training data is not available. However, as opposed to the completely unpaired setting, there exists an indirect link between the training data from different domains. We consider three domains - RGB, Depth, and Semantic segmentation represented as $\mathbf{R}$, $\mathbf{D}$, and $\mathbf{S}$, respectively. We assume we have access to sets of paired training samples from domains $\mathbf{R-D}$ and $\mathbf{R-S}$. These sets are disjointed such that no explicit paired $\mathbf{D-S}$ sample exists. We wish to do inference for the translation between domains $\mathbf{D-S}$. We explore whether the information from the paired sets can be leveraged to obtain better results in this unseen translation.

While this problem can be approached by having independent networks for each domain pair, this formulation requires $N(N-1)/2$ independent models~\cite{anoosheh2018combogan} for $N$ unique domains. It is also not clear with such an approach how paired data from other domains can be leveraged to improve performance for unseen pairs. Wang~\etal~\cite{Wang_2018_CVPR} use domain specific encoders and decoders like in~\cite{anoosheh2018combogan}, and train using only available paired training data. They enforce a latent space consistency loss to ensure encoders from all domains map paired samples to the same space. This helps during inference for unseen paired samples. 

Our aim is to have a single compact network instead of multiple networks with domain specific architectures. Our network consists of only a single encoder and decoder unlike~\cite{Wang_2018_CVPR}. We also believe using a single network might better exploit the structural similarity between the different domains. In order to use a single network for all domains, we need to have a homogeneous representation for all domains in terms of dimensions of their respective samples. Therefore, we represent samples from each domain as 3-channel images. Thus, the single channel depth samples are mapped to 3-channel samples using a standard colormap, and the one-hot encoded semantic segmentation samples are mapped to 3-channel samples using a fixed and unique RGB mapping for each class.

In order to get the image of the target domain as output, we use conditional batch normalization~\cite{de2017modulating} at all layers of the decoder. We explore two different ways of conditioning based on the source input domain $A$, and desired target output domain $B$, and perform a comparison between these approaches. The conditional normalization is done using only a linear embedding. This architectural formulation ensures that there is not a significant increase in network parameters as the number of domains is increased.

\vspace{-3mm}
\paragraph{Encoder $\mathbb{E}$}
The encoder $f(x)$ maps samples from different domains to an unknown latent distribution space. This latent distribution is then used as an input to the decoder.

\vspace{-3mm}
\paragraph{Decoder $\mathbb{G}$}
The decoder $g_{AB}(y)$ takes the latent space output from the encoder as its input and outputs the desired target domain image using domain conditional batch normalization. The mechanism for obtaining the output is explained later in this section.

\vspace{2mm}

We consider paired training data for two sets of domain pairs, namely $\mathbf{R-D}$ and $\mathbf{R-S}$. For the first set, we train the encoder-decoder for translations from $R$ to $D$, and the inverse translation from $D$ to $R$. Similarly, for the second set, we train for translations from $R$ to $S$ and $S$ to $R$. The training loss consists of an adversarial (GAN) loss, an identity loss, and an $L_{1}$ reconstruction loss, and can be written as:

\vspace{-2mm}
\begin{equation}
\begin{aligned}
    \mathcal{L} = \mathcal{L_{GAN}} +  \lambda_{L_{1}}\mathcal{L}_{L_{1}} +  \lambda_{idt}\mathcal{L}_{idt}  
    \label{eq:total_loss}
\end{aligned}
\end{equation}

We use the relativistic least squares GAN loss~\cite{jolicoeur2018ragan} in our model. The GAN loss ensures that the generated output image looks realistic.

The $L_{1}$ reconstruction loss minimizes the $L_{1}$ distance between the translated target image and the corresponding ground truth image. For the $R-D$ domain pair training set, it is defined as:

\vspace{-4mm}
\begin{equation}
\begin{aligned}
    \mathcal{L}_{L_{1}} = \lambda_{R} * \mathbb{E}||(g_{DR}(f(d)) - r||_{1} \\
            + \lambda_{D} * \mathbb{E}||(g_{RD}(f(r)) - d||_{1}
    \label{eq:l1_loss}
\end{aligned}
\end{equation}
where $\lambda_{R}$ and $\lambda_{D}$ are scaling factors for domains $R$ and $D$, respectively.

\vspace{2mm}
The identity loss is the $L_{1}$ distance between an image from a source domain with a translated image where the target domain is same as the source domain. This loss acts as a regularizer for the network. For the $R-D$ domain pair training set, it is defined as:

\vspace{-2mm}
\begin{equation}
\begin{aligned}
    \mathcal{L}_{L_{idt}} = \lambda_{R} * \mathbb{E}||(g_{RR}(f(r)) - r||_{1} \\
            + \lambda_{D} * \mathbb{E}||(g_{DD}(f(d)) - d||_{1}
    \label{eq:idt_loss}
\end{aligned}
\end{equation}

Similar losses can be defined for domain pair training set $R-S$. Training is done by alternating between samples between the two domain pair sets and the networks are updated with the corresponding losses.



We now present the two different approaches of domain normalization to obtain realistic images for the desired target domains.

\subsection{Output Domain Conditional Normalization with Latent Space Invariance (DCN-0)}

In this approach, the decoder normalization is conditioned on the desired output domain. In order for this approach to work, the decoder must learn to map the latent space representation into any of the target domains. We therefore propose that the latent space representation be domain invariant. The decoder will thus take as input the domain invariant latent space and produce a target image through normalization conditioned on the target domain. 

\begin{figure}
\centering
\begin{minipage}[t]{.45\textwidth}
   \includegraphics[width=\linewidth]{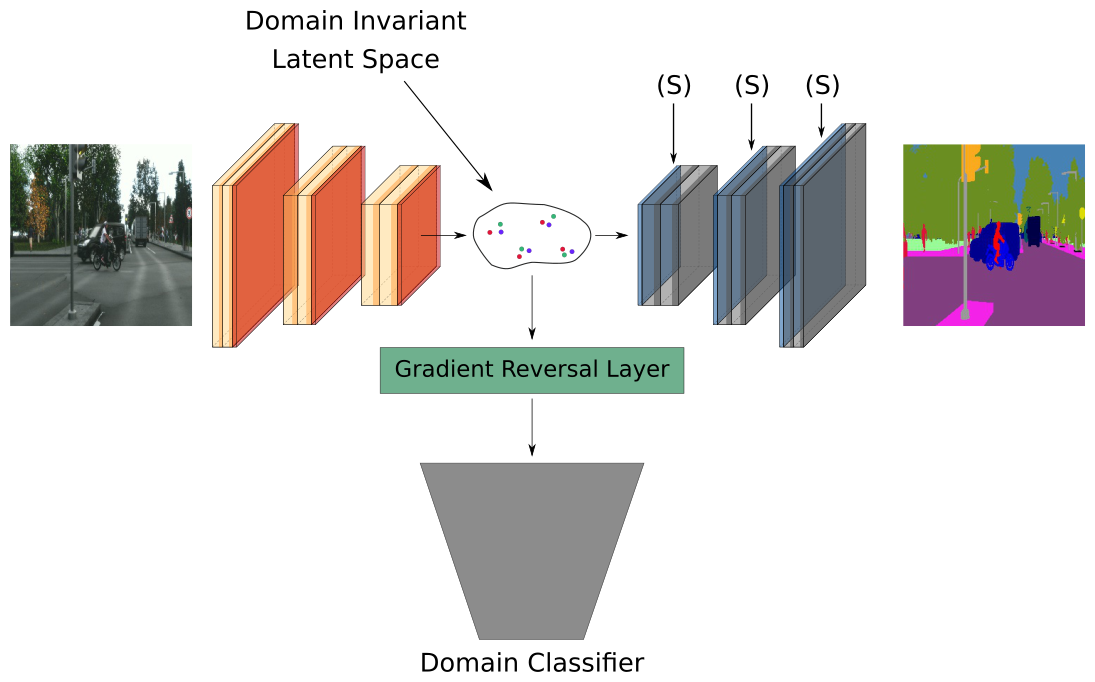}
   \caption{DCN-0: A domain classifier is used which tries to correctly classify the domain of the input source image. The encoder is trained by reversing the gradient from the domain classifier, and thus, it learns to map images from all domains into a domain-invariant latent space.}
\label{fig:approach_1_detailed}
\end{minipage}\qquad
\begin{minipage}[t]{.45\textwidth}

   \includegraphics[width=\linewidth]{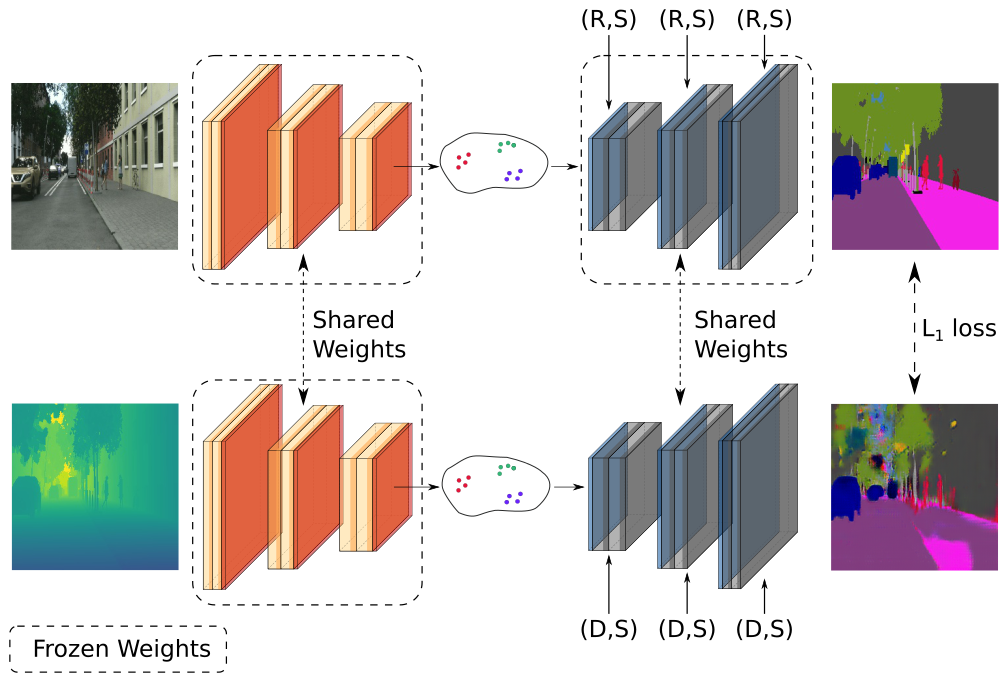}
   \caption{DCN: Learning for unseen domain pair, D-S through pseudo pairs. A paired sample from domains R-D is selected and the corresponding semantic image is obtained for both domains. The loss between the two semantic images is computed to train the decoder.}
\label{fig:approach_2_detailed}

\end{minipage}
\end{figure}

Our objective is to produce the same latent distribution for different domains that represent the same content. For this goal, we explicitly enforce domain invariance using an adversarial loss. We use a domain classifier, which takes the latent space output of the encoder as its input. The objective of this domain classifier is to correctly classify the domain of the sample image. The task of the encoder is to ensure this domain classifier does not make correct classifications. The encoder is thus trained by reversing the gradient obtained from the domain classifier through a gradient reversal layer~\cite{ganin2014unsupervised}, and a scaling factor $\lambda_{cls}$. A visual description is shown in Figure~\ref{fig:approach_1_detailed}.

A potential downside for this domain invariant feature representations is that the resulting performance might be hampered for some domains. If we consider the domains of RGB, semantics, and depth, the domain of RGB is richer in terms of the information it contains with respect to the other two domains. Thus, ensuring domain invariance at the level of latent space representation may possibly result in some loss of information that is required for a successful RGB translation. Nevertheless, it is also important to mention that because of our domain invariant encoder, our decoder can perform translations between domains that has few or nonexistent interactions during training.

\vspace{-1mm}

\subsection{Input-Output Domain Conditional Normalization (DCN)}

In this approach, there is no explicit enforcement of a domain invariant latent space, and the decoder normalization is conditioned on both the input and output domains. If we were to consider domains $R$, $S$, and $D$ as described earlier, the conditional normalization for all possible image translations will be the set $\{$$RR$, $RS$, $RD$, $SR$, $SS$, $SD$, $DR$, $DS$, $DD$$\}$. Since the underlying decoder architecture is the same and the increased size of domain conditional normalization set as compared to DCN-0 (9 vs 3) only affects the normalization layers, this formulation does not increase the total number of parameters significantly, as shown later in Table~\ref{tab:results}.

When training in the zero-paired setting where paired training samples comprise of only sets $R-D$ and $R-S$, no sample exists for the domain pairs $D-S$. Thus, there exists no training data which can use the $DS$ or $SD$ conditional normalization setting. Directly using a poorly trained network for inference of these domain pairs will result in images lacking good quality. In order to rectify this problem, we propose to exploit the already available paired data information. To get a good $D-S$ mapping, we consider a paired training sample for the domain pair $R-D$. We translate this sample to obtain mappings corresponding to $R-S$ and $D-S$. The two mappings obtained should ideally be the same and are referred to as pseudo pairs~\cite{wang2019mix}. Therefore, we can compute the loss between these two images and use it to train the model. However, instead of improving the $D-S$ mapping, such an approach might worsen the $R-S$ mapping. Therefore, we freeze all the weights, expect for the decoder for the $D-S$ mapping. The detailed approach is shown in Figure~\ref{fig:approach_2_detailed}.

\section{Experiments and Results}
\label{sec:experiments}

\subsection{Dataset}
\label{ssc:dataset}

\begin{figure}[t]
\begin{center}
   \includegraphics[width=0.95\linewidth]{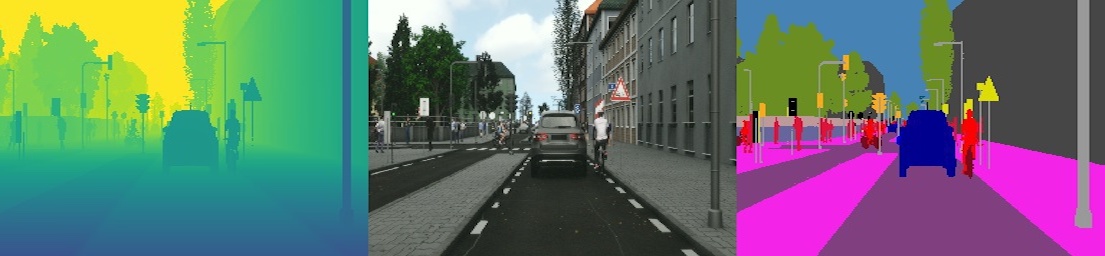}
\end{center}
\vspace{-4.0mm}
\caption{A sample from the Synscapes dataset~\cite{wrenninge2018synscapes} showing the corresponding 3 channel depth, RGB, and semantic segmentation images.}
\label{fig:synscapes_overview}
\vspace{-3.0mm}
\end{figure}

In our experiments, we use the Synscapes dataset~\cite{wrenninge2018synscapes}, which consists of 25,000 synthetically generated RGB images from virtual road scenes. Each RGB image has its corresponding depth and segmentation information. An overview of a sample from the dataset is shown in Figure~\ref{fig:synscapes_overview}. For the zero-pair setting, we only access disjoint subsets of the dataset which comprise of 12,000 samples of either RGB-depth pairs, or RGB-semantic pairs. We also produce a separate disjoint subset of 1000 samples for the test set with images from the depth and semantics domain. As for performance quantitative measures for the generated output results, we are reporting the standard Intersection-over-Union (IoU) and pixel accuracy [\%] with respect to the ground truth semantic labels.

\subsection{Implementation Details}
\label{ssc:implementation_details}

We use the network architecture described in CycleGAN~\cite{zhu2017cyclegan} for all our methods. The network is split into an encoder and decoder. We use 9 ResNet~\cite{He_2016_CVPR} blocks, and consider the output of the 4th block as the latent space output of the encoder. The remaining 5 blocks, along with succeeding components form the decoder. 

Our models are trained on downsampled images of resolution $256\times256$. We conduct the evaluation for the depth to semantic segmentation task. During the evaluation, the segmentation map from these images is obtained using a nearest-neighbor search for each pixel by finding its distance in the RGB color space with all the semantic labels, and assigning it the label with which the distance is minimum. In order to compute the metrics, the segmentation map is upscaled to the original resolution of $720\times1440$ using nearest neighbor upsampling. All models are trained for 240k iterations, with a learning rate which is 0.0002 for the first half of training and linearly decays to 0 for the latter half.

\subsection{Ablation Study}
\label{ssc:ablation_study}

We conducted an ablation study where we see the effect of the individual components and other losses on the overall results for the zero-pair translation approaches. The results are shown in Table~\ref{tab:ablation}:

\begin{table*}[ht]
\setlength{\tabcolsep}{3pt} 
\centering
\resizebox{\linewidth}{!}
{
\begin{tabular}[\linewidth]{|c||c c c c c c c c c c c c c c c c c c c| c| c|}
\hline
&\multicolumn{17}{c}{Intersection-over-Union (IoU)}&&&&\\
Method & \rotatebox{90}{Road} & \rotatebox{90}{Sidewalk} & \rotatebox{90}{Building} & \rotatebox{90}{Wall} & \rotatebox{90}{Fence} & \rotatebox{90}{Pole} & \rotatebox{90}{Traffic Light } & \rotatebox{90}{Traffic Sign} & \rotatebox{90}{Vegetation} & \rotatebox{90}{Terrain} & \rotatebox{90}{Sky} & \rotatebox{90}{Person} & \rotatebox{90}{Rider} & \rotatebox{90}{Car} & \rotatebox{90}{Truck} & \rotatebox{90}{Bus} & \rotatebox{90}{Train} & \rotatebox{90}{Motorcycle} & \rotatebox{90}{Bicycle} & \rotatebox{90}{mIoU} & \rotatebox{90}{Pixel Acc.}\\
\hline\hline
DCN-0 & \textbf{90.8} & \textbf{70.8} & \textbf{82.1} & \textbf{27.8} & \textbf{8.19} & \textbf{19.6} & \textbf{26.8} & \textbf{25.2} & \textbf{76.6} & \textbf{20.1} & 91.5 & \textbf{45.8} & \textbf{20.4} & \textbf{78.1} & \textbf{17.3} & \textbf{28.1} & \textbf{18.1} & \textbf{16.5} & \textbf{13.2} & \textbf{40.9} & \textbf{84.5}\\
DCN-0 w/o domain invariance &  70.1 & 13.7 & 45.9 & 4.89 & 0.3 & 0.6 & 0.1 & 0.0 & 38.4 & 0.0 & \textbf{92.9} & 9.36 & 0.3 & 22.7 & 2.1 & 6.1 & 4.9 & 0.8 & 1.5 & 16.5 & 60.2\\
\hline\hline
DCN & \textbf{98.3} & \textbf{84.9} & \textbf{87.7} & \textbf{46.4} & \textbf{20.6} & \textbf{23.7} & \textbf{29.4} & \textbf{38.2} & \textbf{84.5} & \textbf{65.4} & \textbf{95.0} & \textbf{59.0} & \textbf{33.7} & \textbf{85.1} & \textbf{46.6} & \textbf{29.9} & \textbf{37.8} & \textbf{39.7} & \textbf{28.8} & \textbf{54.3} & \textbf{89.7}\\
DCN w/o pseudo pairs &  7.8 & 0.4 & 10.8 & 0.6 & 0.7 & 0.8 & 0.3 & 0.3 & 9.1 & 0.6 & 0.0 & 2.6 & 0.9 & 0.2 & 1.6 & 1.9 & 1.1 & 0.0 & 0.9 & 2.1 & 9.0 \\
DCN w/o weight freezing & 94.3 & 83.6 & 85.8 & 43.5 & 19.2 & 23.2 & 28.9 & 31.8 & 84.2 & 55.8 & 94.4 & 57.6 & 32.6 & 83.8 & 41.5 & 28.7 & 37.3 & 32.1 & 27.3 & 50.1 & 87.8\\
\hline
\end{tabular}
}
\caption{Ablation results for models trained on depth to semantics image-to-image translation task in a zero-paired setting.}
\label{tab:ablation}
\vspace{-1.0mm}
\end{table*}

It can be seen that for DCN-0 model a domain invariant latent space is required and without enforcing it through the domain classification loss, the model performance reduces drastically. This makes sense since the model is conditioned only on the output domain and thus, the model has no way to deal with the unseen pair translation $D-S$. Similarly, for the DCN model, using pseudo pairs ensures that the $D-S$ translation is encountered while training, and the network learns to correctly map between these domains. Without this loss, the model does not use the $DS$ conditional normalization setting during training, and it is used for the first time during inference, resulting in a poorer performance. The results also show the importance of weight freezing. As expected, in the presence of weight learning, the model starts unlearning the mapping for $R-D$ and $R-S$ pairs, and since these mappings are used as reference to learn the $D-S$ mapping, the resulting model performs worse.

\subsection{Comparison with other Methods}
\label{ssc:comparison_methods}

In addition to our proposed approaches to zero-pair I2I translation we report the performance of several methods.
As baselines, we use CycleGAN~\cite{zhu2017cyclegan} which is trained on unpaired depth-semantic data. We also use a cascaded version of two pix2pix~\cite{isola2017pix2pix} models, which are trained using depth-RGB and RGB-semantic pairs, and for translation from depth to semantics, the output from the first model is used as the input for the second model. For reference, as an upper bound, we report the performance of pix2pix~\cite{isola2017pix2pix} trained with paired data in full supervision for depth-semantic translation. We also compare our method with the state of the art method in zero-pair I2I translation, mix-and-match networks (M\&MNet)~\cite{Wang_2018_CVPR}.

We adopt the M\&MNet method of~\cite{Wang_2018_CVPR} to our architecture and consider 3 separate encoders and 3 separate decoders for the domains of RGB, depth and semantic segmentation. We consider the depth and semantic segmentation samples as 3-channel images for a fair comparison, and optimize the other hyperparameters to work best with the dataset. We must emphasize that the difference between our method and~\cite{Wang_2018_CVPR} is that we use a single encoder-decoder network instead of having multiple networks corresponding to each domain. We also enforce the latent space invariance in DCN-0 through an adversarial domain classifier instead of L2 distance between the latent representations of paired samples from different domains in M\&MNet.

\begin{figure*}[t]
\begin{center}
   \includegraphics[width=0.95\linewidth]{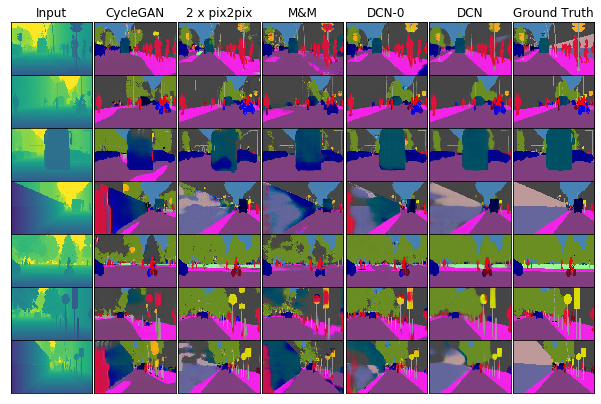}
\end{center}
\vspace{-4.0mm}
\caption{Comparison of results using our method with baseline and other methods for the depth to semantics image-to-image translation task in a zero-paired setting.}
\label{fig:results}
\end{figure*}

\begin{table*}[ht]
\setlength{\tabcolsep}{3pt} 
\centering
\resizebox{\linewidth}{!}
{
\begin{tabular}[\linewidth]{|c||c c c c c c c c c c c c c c c c c c c| c| c| c|}
\hline
&\multicolumn{18}{c}{Intersection-over-Union (IoU)}&&&&\\
Method & \rotatebox{90}{Road} & \rotatebox{90}{Sidewalk} & \rotatebox{90}{Building} & \rotatebox{90}{Wall} & \rotatebox{90}{Fence} & \rotatebox{90}{Pole} & \rotatebox{90}{Traffic Light } & \rotatebox{90}{Traffic Sign} & \rotatebox{90}{Vegetation} & \rotatebox{90}{Terrain} & \rotatebox{90}{Sky} & \rotatebox{90}{Person} & \rotatebox{90}{Rider} & \rotatebox{90}{Car} & \rotatebox{90}{Truck} & \rotatebox{90}{Bus} & \rotatebox{90}{Train} & \rotatebox{90}{Motorcycle} & \rotatebox{90}{Bicycle} & \rotatebox{90}{mIoU} & \rotatebox{90}{Pixel Acc.} & \rotatebox{90}{No. Param.}\\
\hline\hline
\multicolumn{23}{c}{}\\
\multicolumn{23}{c}{\Large \textit{Paired Image to Image Translation, full supervision, upper bound}}\\
\hline
pix2pix ~\cite{Isola_2017_CVPR} & 98.8 & 94.8 & 96.3 & 86.3 & 82.8 & 53.9 & 69.1 & 72.9 & 941. & 81.8 & 96.5 & 81.1 & 64.4 & 93.9 & 79.1 & 71.2 & 66.8 & 62.6 & 73.8 & 69.0 & 93.3 & 22.75M\\
\hline\hline
\multicolumn{23}{c}{}\\
\multicolumn{23}{c}{\Large \textit{Zero-Pair Image to Image Translation}}\\
\hline
CycleGAN ~\cite{zhu2017cyclegan} & 81.2 & 36.8 & 38.7 & 0.51 & 0.28 & 3.54 & 0.07 & 0.18 & 28.6 & 10.9 & 93.1 & 36.9 & 24.0 & 68.8 & 14.9 & 18.6 & 8.91 & 8.95 & 10.1 & 25.5 & 65.2 & 22.75M\\
$2\times$ pix2pix ~\cite{Isola_2017_CVPR} & 93.9 & 81.9 & 83.8 & 36.3 & \textbf{20.9} & 23.0 & 17.1 & \textbf{43.3} & 79.6 & 55.6 & 93.4 & 49.8 & 26.2 & 79.7 & 26.1 & \textbf{31.3} & 26.9 & 35.6 & 27.4 & 48.3 & 87.2 & 45.50M\\
M\&MNet ~\cite{Wang_2018_CVPR} & 80.6 & 28.6 & 71.7 & 14.9 & 0.21 & 24.0 & 22.9 & 3.60 & 50.2 & 0.13 & 93.5 & 38.4 & 16.1 & 63.8 & 13.8 & 15.7 & 9.38 & 4.16 & 2.12 & 29.1 & 75.3 & 27.05M\\
\hline
DCN-0 (ours) & 90.8 & 70.8 & 82.1 & 27.8 & 8.19 & 19.6 & 26.8 & 25.2 & 76.6 & 20.1 & 91.5 & 45.8 & 20.4 & 78.1 & 17.3 & 28.1 & 18.1 & 16.5 & 13.2 & 40.9 & 84.5 & \textbf{11.41M}\\
\textbf{DCN (ours)} & \textbf{98.3} & \textbf{84.9} & \textbf{87.7} & \textbf{46.4} & 20.6 & \textbf{23.7} & \textbf{29.4} & 38.2 & \textbf{84.5} & \textbf{65.4} & \textbf{95.0} & \textbf{59.0} & \textbf{33.7} & \textbf{85.1} & \textbf{46.6} & 29.9 & \textbf{37.8} & \textbf{39.7} & \textbf{28.8} & \textbf{54.3} & \textbf{89.7} & 11.44M\\

\hline
\end{tabular}
}
\caption{Results showing the individual class IoU, mean IoU, pixel accuracy and total number of parameters for different models for depth to semantics image-to-image translation task in a zero-paired setting. }
\label{tab:results}
\vspace{-1.0mm}
\end{table*}


The results are shown in Table~\ref{tab:results} and Figure~\ref{fig:results}. DCN performs the best among the compared zero-pair methods. CycleGAN~\cite{zhu2017cyclegan} method performs as expected since it is trained using purely unpaired data and does not use the paired information for $R-S$ and $R-D$ domain pairs. Another important observation is that the baseline method of $2\times$ pix2pix~\cite{isola2017pix2pix} performs quite well and is very close to the upper baseline of the fully supervised pix2pix model, especially in terms of pixel accuracy (87.2\% vs. 93.3\%). A possible explanation might be the fairly homogeneous synthetic data in this dataset as a result of which the two models learned for $D-R$ and $R-S$ pairs are very good. Thus, the resulting composition of these models performs a $D-S$ mapping with a significantly reduced error. It is noted that both DCN-0, as well as M\&MNet~\cite{Wang_2018_CVPR} perform poorly compared to this baseline. Since both methods rely on obtaining a domain invariant latent space representation, it can be said that the resulting representation in both approaches is not completely invariant to the input domain. DCN, which uses normalization conditioned on both the input and output domains, performs much better while using only a few more additional parameters compared to DCN-0, and still using significantly less parameters than any of the other models. In our experiments, it was also observed that incorporating a domain invariant latent space doesn't improve the performance of DCN. This is understandable since the DCN decoder explicitly learns the mapping between the unseen domain pairs, thus alleviating the need of a domain invariant latent space. We note also that our approaches have $\sim4\times$ fewer parameters than $2\times$ pix2pix, being the lightest solutions.

The difference between our model performance with the upper baseline is significantly smaller in terms of pixel accuracy (89.7\% vs. 93.3\%), as compared to mean IoU (54.3 vs. 69.0). This is primarily because of the different relative representations of different classes and poor performance on some classes which are under-represented in the dataset. Further research is necessary to bridge the performance gap between zero-pair and paired I2I translation methods.

\subsection{Additional results on Scenenet Dataset~\cite{McCormac_2017_ICCV}}

We perform an additional comparison of our method with the M$\&$M method~\cite{Wang_2018_CVPR} on the ScenenetRGBD dataset~\cite{McCormac_2017_ICCV}. The dataset consists of rendered indoor scene videos, from which we use the RGB frames and their corresponding semantic segmentation and depth ground truths. We conduct the experiments on the "51k" dataset used in~\cite{Wang_2018_CVPR}, which uses the first 50 frames from the first 1000 train videos as the training set and the 60th frame from these videos as the test set. For the zero-pair setting, the training set is split into two disjoint sets consisting of RGB-Depth and RGB-Semantics pairs, whereas the inference is done on the unseen Depth-Semantics pairs. The samples are originally of resolution $320\times240$ but are resized to $256\times256$, and converted back to the original size for inference using nearest neighbor interpolation.

We use the Segnet~\cite{badrinarayanan2017segnet} encoder-decoder architecture used in~\cite{Wang_2018_CVPR}, instead of the CycleGAN~\cite{zhu2017cyclegan} architecture used in our earlier experiments. The domain conditional normalization is added to the Segnet architecture, in order to use a single network which handles all domains. For a fair comparison among methods, we modify the network used in~\cite{Wang_2018_CVPR} to use 3-channel images for both depth and segmentation map. We optimize the parameters for their architecture in this modified setting. As an upper baseline, we train a fully supervised pix2pix~\cite{Isola_2017_CVPR} model for the Depth to Semantics task which also uses 3-channel images. The results are shown in Table~\ref{tab:results_scenenet}. We observe that the upper baseline itself has a very low mIoU of 22.5. Nevertheless, the results show that our model performs better than M$\&$M net~\cite{Wang_2018_CVPR}.

\begin{table}[t]
\setlength{\tabcolsep}{8pt} 
\centering
\begin{tabular}{|c||c|c|}
\hline
Method & Pixel Acc. & mIoU \\
\hline\hline
\hline
pix2pix\footnotemark~\cite{Isola_2017_CVPR} & 68.7 & 22.5\\
\hline\hline
\hline
M\&MNet~\cite{Wang_2018_CVPR} & 39.5 & 7.1\\
\hline
\textbf{DCN (ours)}  & \textbf{59.0} & \textbf{10.7} \\
\hline
\end{tabular}

\caption{Results showing the pixel accuracy and mean IoU for different models for depth to semantics image-to-image translation task in a zero-paired setting.}
\label{tab:results_scenenet}
\vspace{-4.0mm}
\end{table}

\footnotetext{Paired I2I, full supervision, serves as upper bound performance.}


\section{Conclusion}
\label{sec:conclusion}

In this paper, we proposed two approaches for zero-pair image-to-image translation which uses only one encoder-decoder network along with domain conditional normalization. Through our experiments, we demonstrated that such a formulation leads to better performance while requiring fewer parameters compared to an approach which uses domain specific encoders and decoders. We also observed that having a domain invariant latent space is difficult to achieve, as seen in DCN-0. Instead, a combined input-output conditional normalization can be used, which leads to notably improved results while requiring only a few more additional parameters, as shown in DCN.
\vspace{0.3cm}

\noindent\textbf{Acknowledgments. }
This work was partly supported by the ETH Z\"urich Fund (OK), a Huawei Technologies Oy (Finland) project, an Amazon AWS grant, and an Nvidia hardware grant.


{\small
\bibliographystyle{ieee_fullname}
\bibliography{egbib}
}

\end{document}